\documentclass{article}

\usepackage{ijcai19}

\usepackage{multirow,multicol}
\usepackage{amsmath}
\usepackage{tabularx}
\usepackage{makecell}
\usepackage{algorithm}
\usepackage{subfig}
\usepackage[noend]{algpseudocode}
\usepackage{url} 

\usepackage{times}
\usepackage{helvet}
\usepackage{courier}
\usepackage{bm}
\usepackage{amsmath}
\usepackage{mdwlist}
\usepackage[noend]{algpseudocode}
\usepackage{relsize}
\usepackage{amsfonts}
\usepackage{graphicx}
\usepackage{caption}
\usepackage{hhline}
\usepackage{color}
\usepackage{colortbl}
\usepackage{booktabs}
\usepackage{textcomp}

\newcommand{\rbm}[1]{\bm{\mathrm{#1}}}

\makeatletter

\def\BState{\State\hskip-\ALG@thistlm}
\makeatother

\title{Learning Task-specific Representation for Novel Words in Sequence Labeling}

\author{
Minlong Peng
\and
Qi Zhang\and
Xiaoyu Xing\and
Tao Gui \and
Jinlan Fu\and
Xuanjing Huang
\affiliations
Shanghai Key Laboratory of Intelligent Information Processing, Fudan University\\
School of Computer Science, Fudan University\\
825 Zhangheng Road, Shanghai, China\\
\emails
\{mlpeng16, qz, xyxing18, tgui16, fujl16, xjhuang\}@fudan.edu.cn
}

\begin{document}

\maketitle

\begin{abstract}

Word representation is a key component in neural-network-based sequence labeling systems. However, representations of unseen or rare words trained on the end task are usually poor for appreciable performance. This is commonly referred to as the out-of-vocabulary (OOV) problem. In this work, we address the OOV problem in sequence labeling using only training data of the task. To this end, we propose a novel method to predict representations for OOV words from their surface-forms (e.g., character sequence) and contexts. The method is specifically designed to avoid the error propagation problem suffered by existing approaches in the same paradigm. To evaluate its effectiveness, we performed extensive empirical studies on four part-of-speech tagging (POS) tasks and four named entity recognition (NER) tasks. Experimental results show that the proposed method can achieve better or competitive performance on the OOV problem compared with existing state-of-the-art methods. 

\end{abstract}

\section{Introduction}

Word representation (or embedding) is a foundational aspect in many state-of-the-art sequence labeling systems \cite{chen2015long,ma2016end,zhang2018chinese}. However, natural language yields a Zipfian distribution \cite{zipf1949human} over words. This means that a significant number of words (in the long tail) are rare. To control model size, a sequence labeling system often constrains the training vocabulary to only cover the top $N$ frequent words within the training set. Those words not covered by the training vocabulary are called OOV words. Learning representations for OOV words is challenged since the standard end-to-end supervised learning methods require multiple occurrences of each word for better generalization. Many works \cite{ma2016end,madhyastha2016mapping} have proved that performance on sequence labeling usually drops a lot when encountering OOV words. This is commonly referred to as the OOV problem, which we are to address in this work.

Over the last few years, many methods have been proposed to deal with the OOV problem. 
These approaches can be roughly divided into two categories: $(1)$ pretraining word representations on very large corpora of raw text \cite{mikolov2013distributed,pennington2014glove,peters2018deep}; $(2)$ further exploiting training data of the task. In a data-rich domain, the first category of methods can often bring considerable improvement to the models that are trained with random initialized word vectors \cite{devlin2018bert}. However, the approach can be criticized when encounter the extremely data-hungry situation. Obtaining sufficient data may be difficult for low-resource domains, e.g. in technical domains and bio/medical domains \cite{deleger2016overview}. 

Therefore, in this work, we highlight methods of the second category. A popular practice of this category is to represent all OOV words with a single shared embedding, which is trained on low-frequency words within the training set, and then assigned to all OOV words at testing time. However, this essentially heuristic solution is inelegant, as it conflates many words thus losing specific information of the OOV words. Another popular practice is to obtain the word representation from its surface-form (e.g., character sequence) \cite{ling2015finding}. This practice is successful at capturing the semantics of morphological derivations (e.g. "running" from "run") but puts significant pressure on the encoder to capture semantic distinctions amongst syntactically similar but semantically unrelated words (e.g. "run" vs. "rung"). Therefore, most state-of-the-art sequence labeling systems will combine the surface-form representation with an unique embedding to represent the word. This again introduces the OOV problem to the systems. 

Recently, a new paradigm of the second direction, which we refer to as the teacher-student paradigm, are being studied. Methods of this paradigm address the OOV problem in two steps. In the first step, they train a supervised model (also called the teacher network in this work) to perform label prediction for those within-vocabulary words to obtain their task-specific word representations. In the second step, they train (or heuristically construct) a predicting model (also called the student network) to predict the representation of a word from its surface-form \cite{pinter2017mimicking}, context \cite{lazaridou2017multimodal}, or the both \cite{schick2018learning}. The training object of the student network is usually to reconstruct representations of those within-vocabulary words. At testing time, when encountering OOV words within a sentence, they first use the student network to predict representations for those OOV words. Then, based on the generated OOV representations, they use the teacher network to perform label prediction. 

Intrinsically, methods of the teacher-student paradigm can be seen as pipelines, with the teacher and student networks being two cascaded components of the pipeline. Though methods of this paradigm has achieved notable success on several tasks, they suffer from the typical error propagation problem of the pipeline paradigm \cite{caselli2015s,bojarski2016end}, since the auxiliary reconstruction object used for training the student network is not guaranteed to be fully compatible with the supervised object used for training the teacher network.

\begin{figure*}
\centering
\includegraphics[width=1.5\columnwidth]{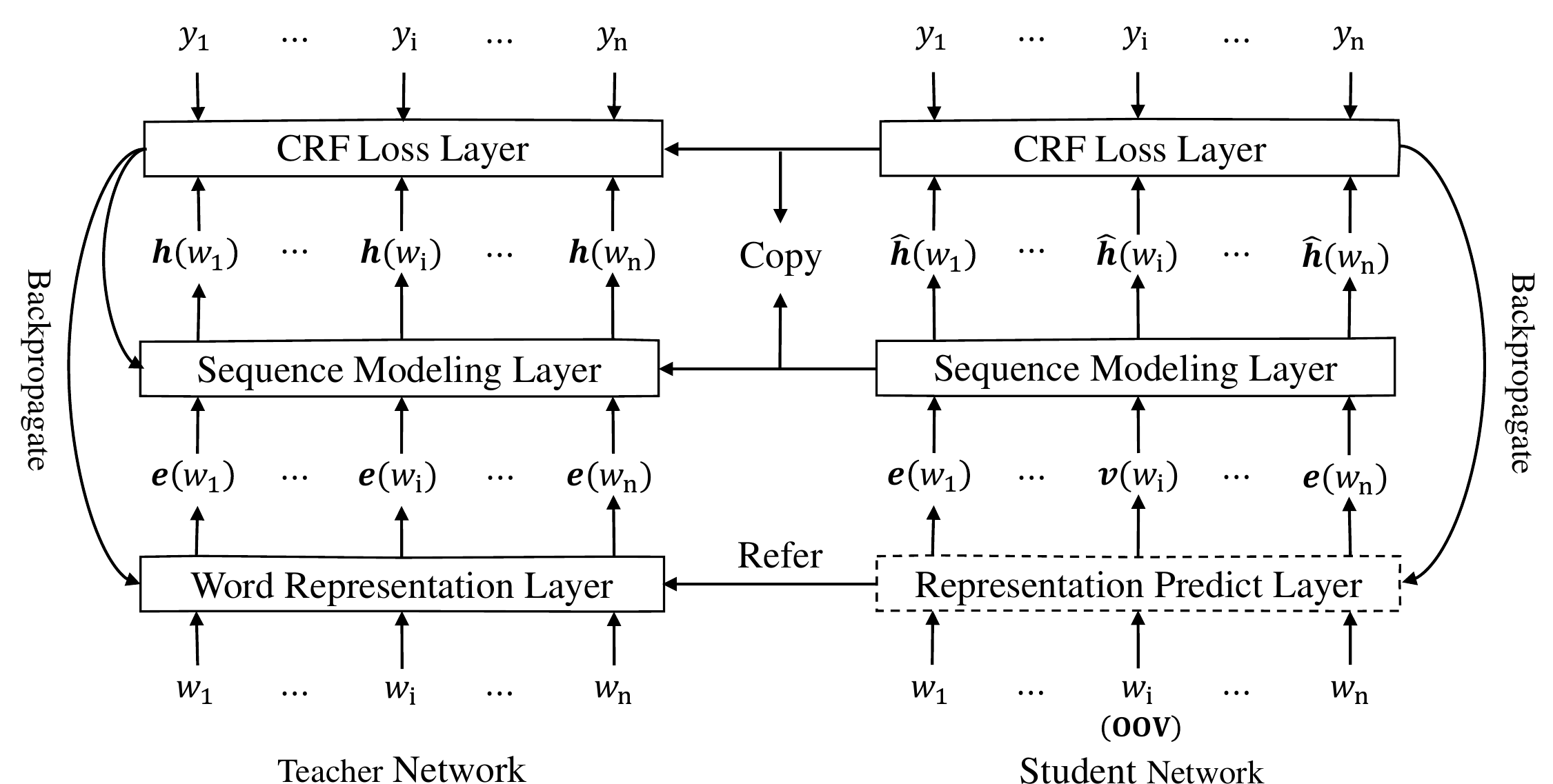}
\caption{Schema of the proposed method when it trains the student network on $w_i$. Boxes with a solid line denote learnable layers of the teacher network and that with a dot line denotes the trainable layer of the student network.}
\label{fig:model}
\end{figure*}

In this work, we propose a novel method of the teacher-student paradigm, which is specifically designed to address the error propagation problem. The main difference of this method from existing ones is the training of the student network. Instead of reconstructing representations of those within-vocabulary words, we train the student network to predict word representations that can achieve good performance on the supervised task. Training signal of the student network is directly backpropagated by the parameter-fixed teacher network. This way, we connect learning of the student network with the teacher network, thus avoiding the error propagation from the prediction of word representation using the student network to the label prediction using the teacher network. We explore applicability of this method to eight sequence labeling tasks in part-of-speech tagging (POS) and named entity recognition (NER). Empirically studies on those tasks show that our proposed method can achieve better or comparative performance on OOV words over existing methods in this paradigm. 

Contributions of this work can be summarized as follows: \textbf{($\bm{i}$)} We propose a novel method of the teacher-student paradigm to address the OOV problem in sequence labeling using only training data of the task. It can avoid the error propagation problem suffered by existing approaches in the same paradigm. 
\textbf{($\bm{ii}$)}  We performed experimental studies on eight part-of-speech tagging and named entity recognition tasks, and achieved better or comparative performance on OOV words over several existing state-of-the-art methods. Source code of this work is available at \url{https://github.com/v-mipeng/TaskOOV}.

\section{Typical Methods of the Teacher-Student Framework}
In this section, we highlight some typical methods of the teacher-student framework.
The first representative work of this framework was proposed by Lazaridou et al., \shortcite{lazaridou2017multimodal}. In that work, they proposed to obtain the representation for an OOV word $w$ through summation over representations of within-vocabulary words occurring in its contexts $\mathcal{C}(w)$:
\begin{equation} \label{eq:mean_pool}
\bm{v}_w(w) = \frac{1}{c(w)} \sum_{C \in \mathcal{C}(w)} \sum_{w^\prime \in C \cap \mathcal{V}} \bm{e}_w(w^\prime).
\end{equation}
Here, $\mathcal{C}(w)$ denotes the set of contexts that contain $w$, $\mathcal{V}$ is the training vocabulary, $c(w)=\sum_{C \in \mathcal{C}(w)} |C \cap \mathcal{V}|$ is the total number of within-vocabulary words in $\mathcal{C}(w)$, and $\bm{e}_w(\cdot)$ is the embedding function defined on within-vocabulary words.

Following this idea, Khodak et al., \shortcite{khodak2018carte} further applied a linear transformation $\rbm{A}$ to the resulting embedding:
\begin{equation} \label{eq:linear}
\hat{\bm{v}}_w(w) = \rbm{A} \bm{v}_w(w),
\end{equation}
to get the representation of $w$. To determine the form of $\rbm{A}$, they trained it on those within-vocabulary words with the training object being to minimize the reconstruction error of an randomly selected within-vocabulary word $w$:
\begin{equation}
\mathcal{L}(w) = d\left(\bm{e}_w(w), \hat{\bm{v}}_w(w)\right),
\end{equation}
where $d(\cdot, \cdot)$ defines the distance between two embeddings, e.g., Euclidean distance function.

In this line, Schick and Schutze, \shortcite{schick2018learning} proposed to model both the context and surface-form (subword n-grams) of OOV words to get their representations:
\begin{equation}
\tilde{\bm{v}}_w(w) = \alpha \hat{\bm{v}}_w(w) + (1-\alpha)\bm{v}_{c}(w),
\end{equation}
where $\bm{v}_{c}(w)$ is a word representation modeling from the surface of $w$, and $\alpha \in [0, 1]$ is a single learnable parameter or a gate generated from $\hat{\bm{v}}_w(w)$ and $\bm{v}_{c}(w)$ as illustrated in \textsection \ref{sec:combine}. 

In general, these methods differ from our proposed method in two points. First, they represent an OOV word $w$ in different contexts with a consistent representation, while we relax the representation of the same word to be different in different contexts. Second, they train the student network on the auxiliary reconstruction object, while we directly train it on the supervised object used for training the teacher network. Of course, the first difference can be easily addressed by adjusting Eq. (\ref{eq:mean_pool}) to:
\begin{equation} \label{eq:base_mean}
    \bm{v}_w(w|C(w)) = \frac{1}{|C(w)|} \sum_{w^\prime \in C(w) \cap \mathcal{V}} \bm{e}_w(w^\prime),
\end{equation}
where $C(w)$ denotes a single context of $w$. According to our experience, this adjustment can generally improve performance of the method in sequence labeling. Therefore, we applied this adjustment to all compared methods of the teacher-student paradigm. Thus, the main difference between our proposed method and existing works is the training strategy of the student network.

\section{Methodology} \label{sec:method}
Figure \ref{fig:model} shows the schema of our proposed method. Generally speaking, the method contains a teacher and a student network. The teacher network is first trained to perform the supervised task. Then, the student network (i.e., the representation predict layer) is trained to generate appropriate representations for words from their contexts and surface-forms. The quality of the generated word representation $\bm{v}(w)$ is measured by the parameter-fixed teacher network (the copied sequence modeling and CRF loss layers), i.e., the training signal of the student network is backpropagated by the teacher network. At testing time, it first uses the student network to generate representations for OOV words and then uses the teacher network to perform label prediction.

\subsection{Notation}
Throughout this work, we denote $\Omega$ the word space and $\mathcal{V}$ the training vocabulary. A sentence consisting of a sequence of words is denoted as $\rbm{X}=\{w_1, \cdots, w_n\} \in \Omega$ and its corresponding label sequence is denoted as $\rbm{Y} = \{y_1, \cdots, y_n\} \in \mathcal{Y}$. The context of a word $w_i$ given its corresponding sentence $\rbm{X}$ is denoted as $\rbm{X}_{\backslash i} = \{w_1, \cdots, w_{i-1}, w_{i+1}, \cdots, w_n\}$, and $\rbm{X}_{<i, j>}$ refers to the sub-sequence $\{w_i, \cdots, w_j\}$. The teacher network is denoted as $T$ and the student network is denoted as $S$.
In addition, let $f$ be a function defined on $\Omega$, we denote $f(\rbm{X})$ a shorthand of $\{f(w_1), \cdots, f(w_n)\}$. 

\subsection{Train the Teacher Network}

The teacher network is used to perform label prediction in the sequence labeling system. It can be of arbitrary architecture suitable for the task. Because the main focus of this work is to deal with the OOV problem instead of designing a superior supervised model, in this work, we tried the typical CNN-based and LSTM-based architectures \cite{huang2015bidirectional} to implement the teacher network. Briefly, this architecture represents a word $w$ with the concatenation of an unique dense vector (embedding) $\bm{e}_w(w)$ and a vector $\bm{e}_c(w)$ modeled from its character sequence using a character-level convolutional neural network (CNN) \cite{kim2014convolutional}:
\begin{equation}
    \bm{e}(w) = [\bm{e}_w(w) \oplus \bm{e}_c(w)],
\end{equation}
where $\oplus$ denotes the concatenation operation. The corresponding sentence representation $\bm{e}(\rbm{X})$ is fed as input into a bidirectional long-short term memory network (BiLSTM) \cite{hochreiter1997long} or a three-layer CNN network \cite{kim2014convolutional} with kernel size set to 3 to model context information of each word, obtaining a hidden representation (not the word embedding) $\rbm{h}(w_i|\rbm{X})$ for each word given $\rbm{X}$. On top of the BiLSTM-based or CNN-based sequence modeling layer, it uses a sequential conditional random field (CRF) \cite{lafferty2001conditional} to jointly decode labels for the whole sentence:
\begin{equation}
p(\rbm{Y}|\rbm{X}; \bm{\theta}_T) = \frac{\prod_{t=1}^m \phi_{t}(y_{t-1}, y_t|\rbm{X})}{\sum_{\rbm{Y}^\prime \in \mathcal{Y}} \prod_{t=1}^m \phi_{t}(y^{\prime}_{t-1}, y^{\prime}_t|\rbm{X})}
\end{equation}
where $\phi_{t}(y^\prime, y|\rbm{X})=\exp(\bm{w}^T_{y^\prime, y} \rbm{h}(w_t) + b_{y^\prime, y})$, $\bm{w}_{y^\prime, y}$ and $ b_{y^\prime, y}$ are trainable parameters corresponding to label pair $(y^\prime, y)$, and $\bm{\theta}_T$ denotes the whole parameters of the teacher network. The training loss of the teacher network is then defined by:
\begin{equation}
\mathcal{L}_T = \sum_{i=1}^N \log p(\rbm{Y}_i|\rbm{X}_i; \bm{\theta}_T),
\end{equation}
where $N$ is the training sentence number.
This network is trained on all words occurring in the training set. After that, we fix its parameters during the training of the student network described in the following.

\subsection{Train the Student Network}
The student network models both surface-form and context information of a word for generating its representation. These two information resources have been demonstrated to be complementary to each other by \citeauthor{schick2018learning}, \citeyear{schick2018learning}.

\paragraph{Model Surface-form.}
The surface-form representation of a word $w$ is obtained from its character sequence $w = \{c_1, c_2, \cdots, c_n\} \in \mathcal{V}_c$, where $\mathcal{V}_c$ is the character vocabulary. Specifically, following the work of \cite{schick2018learning}, we first pad the character sequence with special start and end characters $c_0 = \langle s \rangle$, $c_{n+1}=\langle e \rangle$ and obtain its up $k$-gram set:
\begin{equation}
    G(w) = \cup_{m=1}^k \cup_{i=0}^{n+2-m} \{c_i, \cdots, c_{i+m-1}\}.
\end{equation}
Then, we define the surface-form embedding of $w$ to be the average of all its up $k$-gram embeddings:
\begin{equation}
    \bm{v}_{\text{form}}(w) =  \frac{1}{|G(w)|}\sum_{g\in G(w)} \rbm{W}_{\text{ngram}} (g),
\end{equation}
where $\rbm{W}_{\text{ngram}}$ denotes an embedding lookup table for n-grams, which are trainable parameters of the student network.

\paragraph{Model Context.}
For modelling the context of word $w_i$, we apply a bidirectional long-short term memory network (BiLSTM) on its context word sequence $\rbm{X}_{\backslash i}$. Here, we consider application of this practice in two different situations. In the first situation, there is only one OOV word within a sentence $\rbm{X}$. While in the second situation, there are multiple OOV words within $\rbm{X}$. 

When there is only one OOV word $w_i$ within a sentence $\rbm{X}$, the task-specific representations for the rest words are all known, and we can directly apply BiLSTM to $\rbm{X}_{\backslash i}$ to predict representation for $w_i$. Specifically, we use the forward LSTM to model the sequence from the beginning of $\rbm{X}$ to $w_{i-1}$:
\begin{equation}
\overrightarrow{\rbm{h}}(w_i| \rbm{X}) = \overrightarrow{\text{LSTM}}(\bm{e}(\rbm{X}_{<1, i-1>}))
\end{equation}
and use the backward LSTM to model the sequence from the end of $\rbm{X}$ to $w_{i+1}$ in a reverse order:
\begin{equation}
\overleftarrow{\rbm{h}}(w_i| \rbm{X}) = \overleftarrow{\text{LSTM}}(\bm{e}(\rbm{X}_{<i+1, n>})).
\end{equation}
The forward and backward hidden representations are concatenated to form the context representation of $w_i$ given $\rbm{X}$:
\begin{equation}
\bm{v}_{\text{context}}(w_i| \rbm{X})=[\overrightarrow{\rbm{h}}(w_i| \rbm{X}) \oplus \overleftarrow{\rbm{h}}(w_i| \rbm{X})],
\end{equation}
where $\oplus$ denotes the concatenation operation.

When there are multiple OOV words within a sentence, for a given OOV word $w_i \in \rbm{X}$, the representations of some other words in $\rbm{X}_{\backslash i}$ are also unknown. Therefore, we cannot directly apply the BiLSTM to $\rbm{X}_{\backslash i}$ to get its predicted representation. To address this problem, we propose to iteratively predict representations for the multiple OOV words within a sentence. Let $\bm{v}^t_{\text{context}}(w_i| \rbm{X})$ denote the predicted representation of $w_i$ given $\rbm{X}$ at the $t^{th}$ iteration with $\bm{v}_{\text{context}}^0(w_i| \rbm{X}) = \rbm{0}$. 
At the $(t+1)^{th}$ iteration, for predicting the representation of $w_i$, we apply the BiLSTM to its $t^{th}$ iteration context $\bm{v}_{\text{context}}^t(\rbm{X}_{\backslash i})$, obtaining $\bm{v}_{\text{context}}^{t+1}(w_i|\rbm{X})$. This process repeats for all OOV words to finish the $(t+1)^{th}$ iteration. The iteration proceeds till $t$ reaches a fix value $K$, obtaining the final predicted representation for each OOV word. According to our experience, it is appropriate to set $K=2$. 

\paragraph{Combine Surface-form and Context.} \label{sec:combine}
We finally combine representations of the surface-form and the context to obtain a joint representation $\bm{v}(w| \rbm{X})$ of $w$. In this work, we follow the idea of \cite{schick2018learning} and combine these two representations with a gate:
\begin{equation} \label{eq:generate_representation}
\bm{v}_w(w|\rbm{X}) = \alpha \bm{v}_{\text{form}}(w) + (1-\alpha)\bm{v}_{\text{context}}(w|\rbm{X})
\end{equation}
where
\begin{equation} 
\alpha = \sigma(\bm{w}^T\left[\bm{v}_{\text{form}}(w) \oplus \bm{v}_{\text{context}}(w|\rbm{X})\right] + b).
\end{equation}
Here, $\sigma$ denotes the sigmoid function. For expression consistency, we denote 
\begin{equation}
    \bm{v}_w(w|\rbm{X}; \bm{\theta}_S) =  \bm{v}_w(w|\rbm{X}),
\end{equation}
where $\bm{\theta}_S$ are those trainable parameters of the student network, which actually corresponds to the representation predict layer depicted in Figure \ref{fig:model}.

\paragraph{Training.} 
Because we use the teacher network to perform label prediction, training object of the student network should be learning to generate representations for OOV words that are suitable for the teacher network. According to the idea of previous works \cite{lazaridou2017multimodal,khodak2018carte,schick2018learning}, we may train the student network to minimize the reconstruction errors of those within-vocabulary words with loss function defined by:
\begin{equation} \label{eq:eq_16}
    \mathcal{L}_{recon} = \frac{1}{N}\sum_{i=1}^N \sum_{w \in \rbm{X}_i \cap \mathcal{V}}  d\left(\bm{v}_w(w|\rbm{X}_i; \bm{\theta}_S), \bm{e}_w(w)\right).
\end{equation}
Here, $d(\cdot, \cdot)$ defines a distance between two embeddings, e.g., Euclidean Distance function.
The biggest problem of this practice is that it may suffer from the error propagation problem, since the auxiliary reconstruction training criteria $\mathcal{L}_{recon}$ is not guaranteed to be compatible with training object of the teacher network. 

In this work, we propose to directly connect training of the student network with the training object of the teacher network, sidestepping designing an appropriate auxiliary training criteria. Without loss of generality, we first consider training of the student network in the situation that there is only one OOV word within a given sentence. To simulate this situation, for a training sentence $\rbm{X}$ and its corresponding label sequence $\rbm{Y}$, we randomly sample a word $w_i$ from $\rbm{X}$. The resultant pair $(w_i, \rbm{X}_{\backslash i}, \rbm{Y})$ forms a training example of the student network. By replacing the $i^{th}$ column of $\bm{e}(\rbm{X})$ with $\bm{v}(w|\rbm{X};\bm{\theta}_S)=[\bm{v}_w(w|\rbm{X}; \bm{\theta}_S) \oplus \bm{e}_c(w)]$, we obtain the input $\bm{v}(\rbm{X})=\{\bm{e}(w_1), \cdots, \bm{v}(w_i|\rbm{X};\bm{\theta}_S), \cdots, \bm{e}(w_n)\}$ of the teacher network. Based on $\bm{v}(\rbm{X})$, we obtain new hidden representation for each word $\rbm{\hat{h}}(w_i|\rbm{X})$ and the task loss is then defined by:
\begin{equation} \label{eq:student_loss}
\mathcal{L}_S(\rbm{X}, \rbm{Y}; \bm{\theta}_S, \bm{\theta}_T) = \log \frac{\prod_{t=1}^m \phi_{t}(y_{t-1}, y_t|\rbm{X})}{\sum_{\rbm{Y}^\prime \in \mathcal{Y}} \prod_{t=1}^m \phi_{t}(y^{\prime}_{t-1}, y^{\prime}_t|\rbm{X})},
\end{equation}
where, this time, $\phi_{t}(y^\prime, y|\rbm{X})=\exp(\bm{w}^T_{y^\prime, y} \rbm{\hat{h}}(w_t) + b_{y^\prime, y})$. Note that the training loss of the student network is the same as that of the teacher network, making sure that their training is compatible. For training the student network, we perform parameter update by:
\begin{equation}
\bm{\theta}_S \leftarrow \bm{\theta}_S - \alpha \frac{\partial \mathcal{L}_S(\rbm{X}, \rbm{Y}; \bm{\theta}_S, \bm{\theta}_T)}{\partial \bm{\theta}_S}.
\end{equation}
During the training of the student network, parameters of the teacher network $\bm{\theta}_T$ is fixed. Algorithm \ref{alg:single_oov} shows the general training process of the student network in the situation that there is only one OOV word within a sentence.

To extend this process to the multiple OOV word situation, we sample multiple words from $\rbm{X}$, e.g., $w_i$ and $w_j$, generating an training example of the student network. The loss defined on this example is similar to that in the one OOV situation and minimized over $\bm{\theta}_S$.

\begin{algorithm}[t] 
\caption{Training of the Student Network} \label{alg:single_oov}
\begin{algorithmic}[1]
\BState \textbf{Input:} the teacher network $T$, training dataset $\mathcal{D}$
\BState \textbf{Result:} the student network $S$
\While {$S$ does not converge}
	\State sample $\rbm{X}= \{{w}_1, \cdots, {w}_n\}$ and its corresponding $\rbm{Y}=\{y_1, \cdots, y_n\}$ from $\mathcal{D}$
	\For{$i \in [1, \cdots, n]$}
		\State Generate $\bm{v}(w_i; X)$ for $w_i \in \rbm{X}$;
		\State $\bm{v}({\rbm{X}}) = \{\bm{e}(w_1), \cdots, \bm{v}(w_i| \rbm{X}), \cdots, \bm{e}(w_n)\}$;
		\State Get $\mathcal{L}_S(\rbm{X}, \rbm{Y}; \bm{\theta}_S, \bm{\theta}_T)$ based on $\bm{v}({\rbm{X}})$ according to Eq. (\ref{eq:student_loss});
		\State Update  $\bm{\theta}_S \leftarrow \bm{\theta}_S - \alpha \frac{\partial \mathcal{L}_S(\rbm{X}, \rbm{Y}; \bm{\theta}_S, \bm{\theta}_T)}{\partial \bm{\theta}_S}$.
	\EndFor
\EndWhile
\end{algorithmic}
\end{algorithm}

\section{Experiments}
To evaluate the effectiveness of our proposed method, we performed experiments on four part-of-speech tagging (POS) tasks and four named entity recognition (NER) tasks. These tasks have varying OOV rates, which is defined by the percentage of testing words occurring less than five times in the training set. These tasks share the same architectures of the teacher and student network as illustrated in \textsection \ref{sec:method}. 

\subsection{Datasets}
\paragraph{{POS}:} For POS, we conducted experiments on: (\textbf{1}) {PTB-English}: the Wall Street Journal portion of the English Penn Treebank dataset \cite{marcus1993building}, (\textbf{2}) {RIT-English}: a dataset created from Tweets in English \cite{derczynski2013twitter}, (\textbf{3}) {GSD-Russian}: the Russian Universal Dependencies Treebank annotated and converted by Google\footnote{https://universaldependencies.org/}, and (\textbf{4}) {RRT-Romanian}: the Romanian UD treebank (called RoRefTrees) \cite{barbu2016romanian}. For PTB-English, we followed the standard splits: sections 2-21 for training, section 22 for validation, and section 23 for testing. For {RIT-English} we followed the split protocol of Gui et al., \shortcite{gui2017part}. While, for UD-Russian and UD-Romanian, we used their given data splits. 

\paragraph{{NER}:} For NER, we performed experiments on: (\textbf{1}) {CoNLL02-Spanish}: the CoNLL2002 Spanish NER Shared Task dataset \cite{Sang2002Introduction}; (\textbf{2}) {CoNLL02-Dutch}: the CoNLL2002 dataset of Dutch language; (\textbf{3}) {Twitter-English}: an English NER dataset created from Tweets \cite{zhang2018adaptive}; and (\textbf{4}) {CoNLL03-German}: the CoNLL2003 NER dataset in German \cite{tjong2003introduction}. These datasets are annotated by four types: PER, LOC, ORG and MISC. For datasets except Twitter-English, we used the official split training set for model training, testa for validating and testb for testing. While for Twitter-English, we followed data splits of Zhang et al., \shortcite{zhang2018adaptive}.

\begin{table}[t]
    \centering
    \resizebox{1\columnwidth}{!}{
    \begin{tabular}{lrrrr}
    \toprule \toprule
         \multirow{2}{*}{Dataset} & \multicolumn{2}{c}{Dev} & \multicolumn{2}{c}{Test} \\ 
          & \#OOV & \multicolumn{1}{c}{OOV Rate} & \#OOV & \multicolumn{1}{c}{OOV Rate}\\ \midrule
          \multicolumn{5}{l}{\textbf{POS}} \\ \toprule
         PTB-English        & 8,392 & 6.37\% & 7,528 & 5.81\% \\
         RIT-English        & 774 & 34.52\% & 760 & 33.17\%\\ 
         GSD-Russian         & 21,323 & 17.96\% & 21,523 & 18.31\%\\ 
         RRT-Romanian   &3,965 & 23.22\% & 3,702 & 22.67\%    \\ \midrule
         \multicolumn{5}{l}{\textbf{NER}} \\ \toprule
         CoNLL02-Spanish & 2,216 &50.91\% & 1,544 & 43.38\%\\
         CoNLL02-Dutch  & 1,819 &69.53\% & 2,564 & 65.05\%\\
         Twitter-English & 1,266 & 79.15\% & 4,131 & 79.13\% \\
         CoNLL03-German & 3,928 & 81.27\% & 2,685 & 73.10\%\\
         \bottomrule \bottomrule
    \end{tabular}
    }
    \caption{Number of OOV words (for POS) and entities (for NER) in the development and testing sets, when treating words occurring less than 5 times in the training set as OOV. An entity is treated as OOV if it contains at least one OOV word.}
    \label{table:OOTV_statistic}
\end{table}

Table \ref{table:OOTV_statistic} reports the statistic results of the OOV problem on the development and testing sets of each dataset. From the table, we can see that the OOV rate varies a lot over different datasets.

\subsection{Compared Methods}
We compared to the following baselines:
\begin{itemize}
\item \textbf{RandomUNK}: This baseline refers to the teacher network trained on all words occurring in the training set. At testing time, it represents words not occurring in the training set with a consistent random vector.

\item \textbf{SingleUNK}: This baseline trains the teacher network on words that occur no less than 5 times in the training set. The other infrequent words and those words not occurring in the training set are all mapped to a single trainable embedding $\bm{e}_{\text{UNK}}$, which is trained during model training.

\item \cite{lazaridou2017multimodal}: This baseline uses RandomUNK as the teacher network. At testing time, it first represents OOV words with their context word representations by mean-pooling as defined in Eq. (\ref{eq:base_mean}) based on RandomUNK and then performs label prediction using RandomUNK.

\item \cite{khodak2018carte}: 
This baseline additionally trains a linear transformer $\rbm{A}$ to transform the mean-pooled representation of \cite{lazaridou2017multimodal} as illustrated by Eq. (\ref{eq:linear}) for predicting representations of OOV words. The training of $\rbm{A}$ is performed on words that occur no less than 5 times in the training set. 

\item \cite{schick2018learning}: This baseline is a variant of the proposed method, but its student network is trained on the reconstruction object as defined in Eq. (\ref{eq:eq_16}). It is also related to the work of \cite{schick2018learning} but using the same architecture of the proposed method to model word context and surface-form. 

\item \cite{akbik2018contextual}: This baseline converts the input sentence into a character sequence. Then, it applies a character language model to the character sequence to get the representation of every word within the sentence. Based on the obtained word representations, it applies a LSTM network to model the word sequence and performs final tag recommendation. To make it compatible with the setting of this work, we did not pre-train the language model on external data and not use pre-trained static word embeddings for this baseline.
\end{itemize}

\begin{table*}[t!]
    \centering
    \renewcommand{\arraystretch}{1}
    \resizebox{\textwidth}{!}{
    \begin{tabular}{@{\extracolsep{2pt}}clcccccccccc}
    \hline \hline
         \multirow{2}{*}{Arch} & \multirow{2}{*}{Model} & \multicolumn{2}{c}{PTB-English} & \multicolumn{2}{c}{RIT-English} &  \multicolumn{2}{c}{GSD-Russian} & \multicolumn{2}{c}{RRT-Romanian} \\ \cline{3-4} \cline{5-6} \cline{7-8} \cline{9-10}
         & & Dev & Test & Dev & Test & Dev & Test & Dev & Test \\  \hline
         \multirow{7}{*}{LSTM} & RandomUNK        					& 85.25 & 85.95 & 63.07 & 61.05 &82.06 & 80.97 &86.73 &87.36 \\
         & SingleUNK        					& 86.90 & 88.78 & 61.37 & 61.97 &85.22 &84.68 &90.11 &89.03 \\
         & \cite{lazaridou2017multimodal}    	& 83.90 & 85.67 & 63.82 & 63.16 &85.22 &84.62 &89.53 &90.14  \\
         & \cite{khodak2018carte}    			& 84.03 & 85.67 & 64.21 & 64.07 &86.23 &85.41 &89.94 &90.11 \\
         & \cite{schick2018learning}     		& 87.62 & 89.37 & 64.73 & 62.24 &86.35 &85.49 &90.34 &90.01 \\
         & \cite{akbik2018contextual}   &87.82 &88.90 &58.01 &59.60 &83.69 &83.93 &88.75 &89.03\\
         & Proposed         					& \textbf{88.68} & \textbf{90.53} & \textbf{66.54} & \textbf{64.87} & \textbf{87.28} & \textbf{86.47} & \textbf{91.64} & \textbf{90.28}\\ 
         \hline 
         \multirow{6}{*}{CNN} & RandomUNK        					&88.10 &88.54 &61.88 &63.42 &85.28 &89.87 &88.31 &89.32 \\
         & SingleUNK        					&87.16 &88.84 &60.85 &59.34 &85.28 &86.17 &87.59 &87.16  \\
         & \cite{lazaridou2017multimodal}    	&89.75 &90.74 &61.49 &63.02 &88.86 &90.06 &89.43 &89.57 \\
         & \cite{khodak2018carte}    			&89.89 &90.84 &61.88 &63.42 &89.18 &90.34 &90.01 &90.08 \\
         & \cite{schick2018learning}     		&89.10 &90.61 &62.79 &62.63 &89.13 &90.08 &90.51 &90.43 \\
         & Proposed         					& \textbf{91.33} & \textbf{91.74} & \textbf{65.82} & \textbf{65.27} & \textbf{90.64} & \textbf{91.52} & \textbf{91.74} & \textbf{91.68}\\
         \hline \hline
    \end{tabular}
    }
    \caption{Model performance on the OOV set for part-of-speech tagging when implementing the sequence modeling layer of the teacher network with LSTM-based and CNN-based architectures.}
    \label{table:pos_result}
\end{table*}

\begin{table*}[t!]
\renewcommand{\arraystretch}{1}
    \centering
    \resizebox{\textwidth}{!}{
    \begin{tabular}{@{\extracolsep{2pt}}clcccccccccc}
    \toprule \toprule
         \multirow{2}{*}{Arch} & \multirow{2}{*}{Model} & \multicolumn{2}{c}{CoNLL02-Spanish} & \multicolumn{2}{c}{CoNLL02-Dutch} &  \multicolumn{2}{c}{Twitter-English} & \multicolumn{2}{c}{CoNLL03-German} \\ \cline{3-4} \cline{5-6} \cline{7-8} \cline{9-10}
         & & Dev & Test & Dev & Test & Dev & Test & Dev & Test \\  \midrule
         \multirow{7}{*}{LSTM} & RandomUNK        					&69.36 &72.06 &64.23 &64.08 & 56.88 & 56.38 &55.92 &56.89 \\
         & SingleUNK        					&68.79 &71.59 &67.83 &66.39 & 56.82 & 56.39 &59.69 &60.16 \\
         & \cite{lazaridou2017multimodal}    	&68.61 &69.08 &65.99 &65.43 & 47.72 & 47.20 &47.87 &49.17 \\
         & \cite{khodak2018carte}    			&68.74 &69.53 &66.34 &65.70 & 48.22 & 47.28 &47.97 &49.33 \\
         & \cite{schick2018learning}     		&70.84 &72.88 &68.88 &67.51 & 59.18 & 57.21 &55.83 &58.42 \\
         & \cite{akbik2018contextual} &61.78 &64.06 &60.49 &62.09 &49.68 &50.22 &55.06 &53.01\\
         & Proposed         					& \textbf{73.91} & \textbf{74.63} & \textbf{70.33} & \textbf{70.12} & \textbf{60.14} & \textbf{58.32} & \textbf{60.55} & \textbf{61.79}\\
         \midrule
         \multirow{6}{*}{CNN} &RandomUNK        					&61.07 &61.61 &53.48 &57.31 &44.82 &43.72 &56.23 &56.94 \\
         & SingleUNK        					&56.87 &58.30 &\textbf{60.68} &\textbf{60.46} &\textbf{57.13} &\textbf{57.34} &62.33 &62.07  \\
         & \cite{lazaridou2017multimodal}    	&54.97 &60.39 &53.73 &56.99 &42.91 &46.99 &43.38 &43.54 \\
         & \cite{khodak2018carte}    			&55.12 &60.41 &54.20 &57.00 &48.21 &47.78 &53.19 &53.50 \\
         & \cite{schick2018learning}     		&61.23 &61.54 &53.60 &57.48&46.79 &46.59 &56.16 &57.44 \\
         & Proposed         					& \textbf{63.38} & \textbf{63.02} & {59.24} & {60.33} & {57.06} & {57.32} & \textbf{62.42} & \textbf{63.01}\\
         \bottomrule \bottomrule
    \end{tabular}
    }
    \caption{Model performance on the OOV set for named entity recognition.}
    \label{table:ner_result}
\end{table*} 

\subsection{Implementation Detail}
For data prepossessing, all digits were replaced with the special token "$<$NUM$>$", and all url were replaced with the special token "$<$URL$>$". Dimension of word embedding, character embedding, and LSTM were respectively set to 50, 16, and 50 for both the teacher and student networks. Kernel size of the character CNN was set to 25 for kernel width 3 and 5. Optimization was performed using the Adam step rule \cite{kinga2015method} with the learning rate set to 1e-3. 

\subsection{Evaluation}
We partitioned the testing (or development) set into two subsets: within-vocabulary (WIV) words and out-of-vocabulary (OOV) words. A word is considered WIV if it occurs more than 5 times in the training set, otherwise OOV. For NER, an entity is considered being of OOV if it contains at least one word of OOV word set. We report model performance (accuracy for POS and F1 for NER) on OOV set. This is because we can exactly tell which subset a word belongs to, thus we can easily combine the best model on each subset to achieve the best overall performance on the whole testing set. For example, we can perform label prediction for OOV words using our proposed model, and perform label prediction for WIV words using the best performing model on the WIV subset.

\subsection{Main Results}

Table \ref{table:pos_result} and \ref{table:ner_result} reports model performance on the OOV set for POS-tagging and NER, respectively, when using BiLSTM and CNN to implement the teacher network. From this table, we have the following observations: (1) on most tasks, methods dealing with the OOV problem outperform the RandomUNK baseline. This verifies the necessity to deal with the OOV problem in sequence labeling. (2) the method \cite{schick2018learning} using both surface-form and context information to generate representations of OOV words outperforms the method \cite{khodak2018carte} using only context information on most datasets. This shows the complementary of surface-form and context information; (3) the most comparative baseline in the teacher-student paradigm \cite{schick2018learning} in general outperforms SingleUNK. This verifies the effectiveness of the motivation beneath the teacher-student paradigm; (4) our proposed method consistently outperforms \cite{schick2018learning}, which differs from the proposed method in the training of the student network. This, on one hand, shows the existence of error propagation from the student network to the teacher network, and on the other hand approves the effectiveness of our proposed method for addressing OOV problem; finally (5) our proposed method consistently outperforms the character language model \cite{akbik2018contextual}. A possible explanation of this result is that our method can use word-level embedding without suffering from the OOV problem while the character language model cannot.

\section{Conclusion}
In this work, we proposed a novel method to address the out-of-vocabulary problem in sequence labeling systems using only training data of the task. It is designed to generate representations for OOV words from their surface-forms and contexts. Moreover, it is designed to avoid the error propagation problem suffered by existing methods in the same paradigm. Extensive experimental studies on POS-tagging (POS) and named entity recognition (NER) show that this method can achieve superior or comparable performance over existing methods on the OOV problem. 
 
\section{Acknowledgements}
The authors wish to thank the anonymous reviewers for their helpful comments. This work was partially funded by China National Key R\&D Program (No. 2018YFC0831105, 2017YFB1002104, 2018YFB1005104), National Natural Science Foundation of China (No. 61532011, 61751201), Shanghai Municipal Science and Technology Major Project (No.2018SHZDZX01), STCSM (No.16JC1420401,17JC1420200), ZJLab. 

\bibliographystyle{ijcai19}
\fontsize{10pt}{10.1pt} \selectfont
\bibliography{TSWE}

\end{document}